\pdfoutput=1

\documentclass[11pt]{article}

\usepackage{ACL2023}

\usepackage{times}
\usepackage{latexsym}

\usepackage[T1]{fontenc}

\usepackage[utf8]{inputenc}

\usepackage{microtype}

\usepackage{inconsolata}

\usepackage{amsmath}
\usepackage{multirow}
\usepackage{listings}
\usepackage{color}
\usepackage{graphicx}
\usepackage{subcaption}
\usepackage{makecell}
\usepackage{booktabs}
\usepackage{cleveref}
\usepackage{colortbl}
\usepackage{soul}
\usepackage{xcolor}
\usepackage{empheq}
\usepackage[many]{tcolorbox}
\usepackage{stfloats}

\definecolor{dkgreen}{rgb}{0,0.6,0}
\definecolor{gray}{rgb}{0.5,0.5,0.5}
\definecolor{mauve}{rgb}{0.58,0,0.82}

\definecolor{dgreen}{rgb}{0.412,0.741,0.271}
\definecolor{dblue}{rgb}{0.220,0.325,0.639}
\definecolor{dred}{rgb}{0.933,0.122,0.137}

\definecolor{g1}{HTML}{b3e2cd}
\definecolor{r1}{HTML}{fdcdac}
\definecolor{w1}{HTML}{cbd5e8}
\definecolor{b1}{HTML}{fff7bc}

\definecolor{lr}{HTML}{bebada}
\definecolor{fr}{HTML}{fccde5}

\definecolor{Lavender}{HTML}{BF94E4}

\newcommand{\ie}{\textit{i}.\textit{e}.\,}
\newcommand{\eg}{\textit{e}.\textit{g}.\,}

\newcommand{\centerone}[2]{\multicolumn{1}{>{\columncolor{#1}}c}{#2}}
\newcommand{\centeroneg}[1]{\centerone{g1}{#1}}
\newcommand{\centeroner}[1]{\centerone{r1}{#1}}

\newcommand{\centeroneb}[1]{\centerone{b1}{#1}}

\definecolor{l1}{RGB}{189,215,238}
\definecolor{l2}{RGB}{222,235,247}
\definecolor{l3}{RGB}{255,230,153}
\definecolor{l4}{RGB}{248,203,173}
\definecolor{l5}{RGB}{244,177,131}

\newtcbox{\columntcbox}{highlight math style={
        colback=gray!30,
        arc=2pt,
        outer arc=2pt,
        boxrule=0pt,
        top=2pt,
        bottom=2pt,
        left=2pt,
        right=2pt,
    }
}

\colorlet{LightLavender}{Lavender!35!}
\newtcbox{\inlinetcbox}[1][]{on line, 
        boxsep=2pt, left=0pt,right=0pt,top=0pt,bottom=0pt,
        colframe=white,colback=LightLavender,  
        highlight math style={enhanced}, #1
}
\newcommand{\inlinetcboxmathcal}[3][]{\inlinetcbox[#1]{$\mathcal{#2}_{#3}$}}

\newcommand{\inlinetcboxmathcalandothers}[4][]{\inlinetcbox[#1]{$\mathcal{#2}_{#3}{#4}$}}

\newcommand\blfootnote[1]{%
  \begingroup
  \renewcommand\thefootnote{}\footnote{#1}%
  \addtocounter{footnote}{-1}%
  \endgroup
}

\title{\texttt{SkillQG}: Learning to Generate Question for Reading \\ Comprehension Assessment}

\author{
    Xiaoqiang Wang\textsuperscript{\rm 1},
    Bang Liu\textsuperscript{\rm 2$\dagger$$\ddagger$},
    Siliang Tang\textsuperscript{\rm 1$\ddagger$} \and
    Lingfei Wu\textsuperscript{\rm 3$\ddagger$} \\
    \textsuperscript{\rm 1}Zhejiang University, \textsuperscript{\rm 2}Universit{\'e} de Montr{\'e}al \& Mila, \textsuperscript{\rm 3}Pinterest \\
    \{\texttt{xq.wang, siliang}\}\texttt{@zju.edu.cn} \\
    \texttt{bang.liu@umontreal.ca, lwu@email.wm.edu} \\
}

\begin{document}
\maketitle
\begin{abstract}

We present \textbf{\texttt{SkillQG}}: a question generation framework with controllable comprehension types for assessing and improving machine reading comprehension models. 
Existing question generation systems widely differentiate questions by \textit{literal} information such as question words and answer types to generate semantically relevant questions for a given context.
However, they rarely consider the \textit{comprehension} nature of questions, \ie the different comprehension capabilities embodied by different questions.
In comparison, our \texttt{SkillQG} is able to tailor a fine-grained assessment and improvement to the capabilities of question answering models built on it.
Specifically, we first frame the comprehension type of questions based on a hierarchical skill-based schema, then formulate \texttt{SkillQG} as a skill-conditioned question generator.
Furthermore, to improve the controllability of generation, we augment the input text with question focus and skill-specific knowledge, which are constructed by iteratively prompting the pre-trained language models.
Empirical results demonstrate that \texttt{SkillQG} outperforms baselines in terms of quality, relevance, and skill-controllability while showing a promising performance boost in downstream question answering task.
\end{abstract}

\blfootnote{$^\dagger$Canada CIFAR AI Chair.}
\blfootnote{$^\ddagger$Corresponding authors.}

\section{Introduction}
\label{sec:introduction}

Question generation (QG) systems aim to generate natural language questions conditioned on a text passage.
As a dual task of question answering (QA), QG is widely applied to create question-answer pairs as data augmentation for QA training~\cite{zhang-bansal-2019-addressing, liu2020asking}, help chatbots continue a conversation with human users~\cite{mostafazadeh-etal-2016-generating, shum2018eliza}, and facilitate reading assessment~\cite{heilman-smith-2010-good, jia2021eqg}.

Most prior QG research has typically focused on generating factoid-based questions that are relevant to a piece of the fact of a single sentence~\cite{zhou2017neural, liu2019learning, zhao-etal-2022-educational}.
Recently, motivated by building the read comprehension (RC) systems that are competent in understanding and reasoning~\cite{kaushik-lipton-2018-much, sinha-etal-2019-clutrr}, there is an increasing interest in developing systems that are capable of generating deep questions~\cite{pan-etal-2020-semantic, fei-etal-2022-cqg}.
However, these works generate diverse questions by relying on different surface-level mentioned information~\cite{cheng-etal-2021-guiding} and consider primarily simple connections between two facts in the context (\eg bridge and intersection). 
Less explored have been more facts and the deeper comprehension types between them~\cite{desai-etal-2018-generating}, such as analysis of discourse relations~\cite{johnstone2017discourse}, a thorough evaluation of stated arguments, and deduction of the high-level semantics~\cite{gao-etal-2022-makes}.
As shown in Figure~\ref{fig:motivation-examples}, $Q_1$ asks for the mentioned facts in stories (\eg ``\textit{The princess climbed out the window of the high tower}''), whereas $Q_2$ and $Q_3$ ask for a deep connection about the events (causal relation in $Q_2$ and future prediction in $Q_3$).

\begin{figure}[!t]
\begin{tcolorbox}
    \small
    \textbf{Context}: \textit{The princess climbed out the window of the high tower and climbed down the south wall when her mother was sleeping. She wandered out a good way. Finally, she went into the forest where there are no electric poles.}
    
   \pmb{$Q_1$}: \textit{Who climbed out of the castle?} \pmb{A}: \textit{Princess.}
    
    \pmb{$Q_2$}: \textit{Why did the princess climb out when her mother was sleeping?} \pmb{A}: \textit{In case of being caught.}
    
   \pmb{$Q_3$}: \textit{What would happen if her mother was not sleeping?} \pmb{A}: \textit{The princess would be caught soon.}
\end{tcolorbox}
\vspace{-3mm}
\caption{Example questions that require different comprehension capabilities to answer.}
\label{fig:motivation-examples}
\vspace{-5mm}
\end{figure}

    
    
    
We argue that generating questions with deeper comprehension brings two major benefits:
(\romannumeral 1) compared with factoid-based QG models, it reflects higher cognitive skills and requires an in-depth understanding of the input text and reasoning over relevant contexts, better imitating how human intelligence embodies the application and integration of skills; (\romannumeral 2) compared with existing deep QG models, it can help build more controllable questions with different comprehension types rather than literal information such as answer types.
Based on such questions, we can better identify the downstream performance of QA systems in specific comprehension types, and assess their corresponding intrinsic ability, further allowing us to provide tailored guidance to them and improve training efficiency.


In this paper, we propose \texttt{SkillQG}: a question generation framework with controllable comprehension types.
Specifically, we define the comprehension types as five skill dimensions ordered by cognitive complexity: \textsc{Remember}, \textsc{Understand}, \textsc{Analyze}, \textsc{Create}, and \textsc{Evaluate}, which are inspired by Bloom's Taxonomy~\cite{krathwohl2002revision}, an educational schema by which teachers structure a curriculum to ensure that learners possess the necessary abilities before progressing to more complex tasks.
Based on the definition, we can better differentiate questions from cognitive demands than previous surface-level information and formulate \texttt{SkillQG} as question generation conditioned on the given comprehension skill.

Furthermore, to improve the specificity of generating questions with a certain comprehension skill, we devise a set of prompts based on the indicative words and question templates of Bloom's Taxonomy.
Using these prompts to iteratively elicit chain-of-thought reasoning of pre-trained language model (PLM), we explicitly generate question focuses (what to ask about) and skill-specific knowledge (how to ask it) to augment the input context.

Finally, to evaluate the \texttt{SkillQG} framework, we introduce evaluation protocols covering question content quality, skill controllability, and downstream QA performance improvement when incorporating the generated questions as additional training data.
Our experimental results show that \texttt{SkillQG} can produce more relevant and skill-controllable questions compared to baseline QG models, and boost the QA performance significantly.

\begin{table*}[t!]
    \resizebox{1.0\textwidth}{!}{
        \begin{tabular}{lll}
            \toprule
            \multicolumn{1}{c}{\textbf{Skill}}& \multicolumn{1}{c}{\textbf{Description}}& \multicolumn{1}{c}{\textbf{Example}} \\
            \hline
            \multirow{2}{*}{\textbf{\textsc{Remember}}}& \cellcolor{l1}& \cellcolor{l1} Factoid: what is X?, when did X happen? \\
            & \multirow{-2}{8cm}{\cellcolor{l1} Retrieve relevant facts from input passage.}& \cellcolor{l1} Definition: what does X mean? \\
            \hline
            \multirow{4}{*}{\textbf{\textsc{Understand}}}& \cellcolor{l2}& \cellcolor{l2} Interpreting: how would you rephrase X? \\
            & \cellcolor{l2}& \cellcolor{l2} Classifying: what is an example of X? \\
            & \cellcolor{l2}& \cellcolor{l2} Summarizing: what is the main idea of X? \\
            & \multirow{-4}{8cm}{\cellcolor{l2} Construct meanings from recalled facts.}& \cellcolor{l2} Comparing: how would you compare X and Y? \\
            \hline
            \multirow{2}{*}{\textbf{\textsc{Analyze}}}& \cellcolor{l3}& \cellcolor{l3} Explanation: what caused X? \\
            & \multirow{-2}{8cm}{\cellcolor{l3} Break facts into its constituent parts and determine how the parts are related to one another.}& \cellcolor{l3} Consequence: what will X cause? \\
            \hline
            \textsc{\textbf{Create}}& \cellcolor{l4} Re-organize elements into a new pattern or structure.& \cellcolor{l4} Predicting: would it arrive on time? \\
            \hline
            \multirow{2}{*}{\textbf{\textsc{Evaluate}}}& \cellcolor{l5}& \cellcolor{l5} Judgment: what do you think of X? \\
            & \multirow{-2}{8cm}{\cellcolor{l5} Make judgments based on established criteria.}& \cellcolor{l5} Justification: why is X the case? \\
            \bottomrule
        \end{tabular}
        }
    \caption{Formulation of hierarchical comprehension skills. Skills are sorted by levels of cognition (lower to higher). See Section~\ref{subsec:formulation-of-comprehension} for details.}
    \vspace{-5mm}
    \label{tab:skill-schema}
\end{table*}

\section{Methodology}
\label{sec:methodology}

In this section, we elaborate our \texttt{SkillQG} for generating skill-infused questions.
Specifically, we first define the comprehension types of questions as a 5-dimensional skill schema, which is drawn upon Bloom's Taxonomy~\cite{krathwohl2002revision} of research in cognitive science and describes the cognitive load of different levels of topics or samples.
Based on this schema, we categorize the questions into different comprehension skills, regarding \texttt{SkillQG} as a conditional generator given a skill.
Furthermore, to improve the controllability of the skill-infused questions, we adapt the indicative words and templates of Bloom's Taxonomy as a set of prompts to discover question focuses and skill-specific knowledge by prompting PLM iteratively.
Finally, these question focuses and knowledge text act as auxiliary inputs to steer the question generator.

\subsection{Formulation of Comprehension Types}
\label{subsec:formulation-of-comprehension}

Question generation has long served as an essential component for knowledge learning~\cite{tobin1990research, lai-etal-2017-race} and assessing learning progress~\cite{holme2003assessment, yudkowsky2019assessment}, especially asking questions about texts at various comprehension levels deepens the understanding of the text and aids in the learner's understanding and growing from what they have read~\cite{holme2003assessment}.
Among relevant research in cognitive science and pedagogy, Bloom's Taxonomy~\cite{krathwohl2002revision} is one of the most basic and influential theories.
Bloom's Taxonomy is a cognition model used for the classification of educational learning objectives into levels of complexity and specificity, including knowledge, comprehension, application, analysis, synthesis, and evaluation.
Inspired by the hierarchical cognitive objectives of Bloom's Taxonomy, we define the comprehension types of questions as a 5-dimensional skill-based schema in Table~\ref{tab:skill-schema}.
We sketch out the meaning of each comprehension skill with some examples as follows.

\noindent
\textbf{\textsc{Remember}.} \
The objective of this skill is to promote retention of the presented material in the same form as it exists.
Therefore, it requires retrieving relevant content from what a model has read, \eg \textit{recall the dates of some events in the input passage}.
Empirically, \citet{sugawara-etal-2018-makes} has shown that some questions can be answered correctly by just string-based matching with the given passage.
In this study, the factoid-based questions~\cite{zhou2017neural} involving a single fact with explicit mentions and definition questions are categorized into this kind of comprehension skill.

\noindent
\textbf{\textsc{Understand}.} \
To build a holistically semantic representation of text from recalled facts in the passage, the easiest way is to build connections between the ``new'' knowledge to be gained and their prior knowledge.
We exemplify four kinds of questions to represent this skill, consisting of interpreting (\eg \textit{paraphrase important speeches and documents}.), classifying (\eg \textit{classify observed or described cases of mental disorders}.), summarizing (\eg \textit{write a brief summary of the events portrayed on a videotape}.), and comparing (\eg \textit{compare historical events to contemporary situations}.).


\noindent
\textbf{\textsc{Analyze}.} \
To step towards a higher comprehension skill, break-down-then-combination is required.
This skill aims to break facts into their constituent parts and determine how the parts are related to one another.
It usually involves the relationships between two events that are causally related where the prior events causally lead to the latter event in question.
Similarly, \citet{ko-etal-2020-inquisitive} reveals that cause-effect analysis is more challenging in understanding tasks than bridging or comparing the known facts, particularly for the cases where the passage contains no explicit causal conjunctions and corresponding background knowledge is required.
Therefore, we include explanation (\eg \textit{why are the stock prices retreating}?) and consequence questions (\eg \textit{what happened to Timmy after he got in the hamper}?) in this skill.

\noindent
\textbf{\textsc{Create}.} \
One of the highest cognitive levels is to put elements together to form a coherent whole.
Although it seems impossible to empower a data-driven model with creative thinking, this skill asks for the possible outcome of a current event, which is predictable based on the existing information in the text.
Inspired by the existing datasets that find textual clues and use them to guess what would happen next~\cite{gao-etal-2022-makes}, we instantiate this comprehension skill as predicting questions (\eg \textit{How will the other animals treat the duckling}?).

\noindent
\textbf{\textsc{Evaluate}.} \
The other of the highest cognitive levels is making judgments based on criteria and standards.
Because the criteria are constructed based on either elaborated details in the passage or external commonsense knowledge, this skill reflects the application of something known into a new scenario.
Besides, this skill helps find out internal inconsistencies and also benefits the development of \textsc{Create} skill.
We classify the judgment (\eg \textit{what do you think of the scientist's conclusions}?) and justification questions into this comprehension skill.


\begin{figure*}[!t]
    \includegraphics[width=\textwidth]{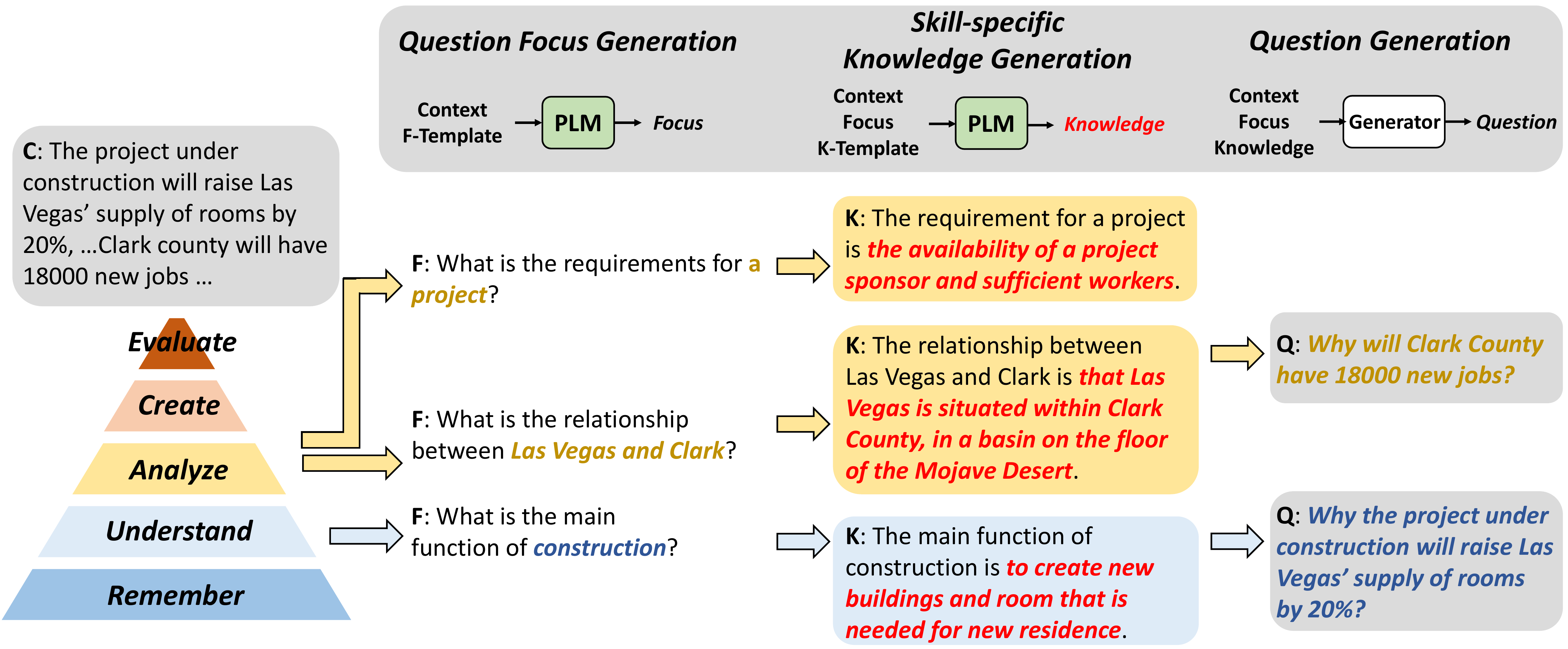}
    \caption{ Illustration of \texttt{SkillQG} pipeline.
    A skill-infused question (\textbf{Q}) is generated from the following steps: question focus generation, skill-specific knowledge generation, and question generation conditioned on the corresponding context (\textbf{C}), question focus (\textbf{F}), and elicited knowledge (\textbf{K}).
    The \textbf{PLM} represents an off-the-shelf GPT2 model, while the \textbf{generator} is initialized from a pre-trained BART model and fine-tuned on the training set.
    }
    \label{fig:pipeline}
    \vspace{-5mm}
\end{figure*}

\subsection{\texttt{SkillQG}}
\label{subsec:skill-qg}

Based on the formulation of comprehension types, we follow the common question generation setup~\cite{zhou2017neural, liu2020asking} and frame \texttt{SkillQG} by a sequence-to-sequence question generator.
Formally, given a context $c$, answer $a$ and comprehension skill $s$, we aim to generate a question $q$ that reflects the corresponding skill by modeling the conditional probability $p_\theta \left(q \mid c, s, a\right)$:
\begin{align}
    p_\theta \left(q \mid c, s, a \right) = \prod_{t=1}^{T}{p_\theta \left(q_t \mid q_{<t}, c, s, a \right)}
    \label{eq:generation}
\end{align}
where $T$ is the length of generated question comprised of a sequence of tokens $q = \langle q_1, \cdots, q_t, \cdots, q_T \rangle$, and the generator is parameterized by $\theta$.
To improve the controllability of generation, we further guide the generator with question-worthy concepts and skill-specific knowledge.
Precisely, we leverage chain-of-thought prompting~\cite{wei2022chain, madaan2022memory} of PLM, a prompting paradigm of successively eliciting relevant knowledge from PLM, to steer the generation of skill-infused questions.
Based on it, we can first capture the question focuses and then externalize the implicit knowledge required for mastering the given comprehension skill.

As illustrated in Figure~\ref{fig:pipeline}, we design several pairs of templates for each comprehension level, \ie \textit{F-template} and \textit{K-template}, denoted as $\mathcal{T}_F$ and $\mathcal{T}_K$ respectively.
These template pairs are with a form of information-seeking questions~\cite{bruner1961act}, such as ``\emph{What is the definition of} \inlinetcbox{\phantom{i}\_\phantom{i}}'' and ``\emph{The definition of} \inlinetcbox{\phantom{i}\_\phantom{i}} \emph{is} \inlinetcbox{\phantom{i}\_\phantom{i}}'', which can help PLM talk with itself to explicitly discover what it cares about when given a comprehension skill.
More specifically, the $\mathcal{T}_F$ together with the input context is used to construct the prompt input for discovering possible question focuses by template-infilling, while the $\mathcal{T}_K$ can generate skill-related knowledge based on the context and question focuses.
Finally, we take the generated knowledge as an auxiliary context and expect that it can contribute to
improving the generation quality.
Denoting the question focus and knowledge text as $f$ and $k$, respectively, the above procedure can be formulated as:
\begin{align}
    f &= \mathcal{M} \big( \mathcal{P}_{F} (c) \big) \label{eq:question-focus-generation} \\
    k &= \mathcal{M} \big( \mathcal{P}_{K} (c, f) \big) \label{eq:knowledge-generation} \\
    c &= \text{Aug}(c, f, k)
\end{align}
where $\mathcal{M}$ denotes the employed PLM, \ie GPT2, $\mathcal{P}_{F}$ and $\mathcal{P}_{K}$ represents the prompt input constructed by $\mathcal{T}_{F}$ and $\mathcal{T}_{K}$, respectively.
$\text{Aug}(c, f, k)$ means augmenting the original context with elicited question focus and knowledge text.

\noindent
\textbf{Question focus generation.} \
To improve the controllability of generated questions, we take inspiration from the chain-of-thought prompting to capture question focuses and skill-related knowledge.
Precisely, considering the close association between the comprehension skill and its involved narrative elements and questioning styles, we devise several pairs of \textit{F-template} $\mathcal{T}_F$ and \textit{K-template} $\mathcal{T}_K$ for each skill. An example is shown in Figure~\ref{fig:pipeline} and all of the templates are summarized in Appendix~\ref{sec:fk-templates}.
They are adapted from the indicative words and question templates of Bloom's Taxonomy.
After that, question focus is generated by feeding the context and $\mathcal{T}_F$ into PLM.
Following the prompt format of causal language models such as GPT2~\cite{radford2019language}, the prompt input $\mathcal{P}_F$ in Eq.~\ref{eq:question-focus-generation} of question focus generation is built as:
\begin{equation*}
    \mathcal{P}_F(c) = \inlinetcbox[boxsep=5pt]{c} \text{\texttt{From the context:}} \inlinetcboxmathcal{T}{F}
\end{equation*}

\noindent
\textbf{Implicit knowledge generation.} \
We further utilize \textit{K-template} $\mathcal{T}_K$ to inquire PLM for generating skill-related knowledge.
This kind of knowledge-externalization method has shown substantial improvements in zero-shot commonsense reasoning~\cite{shwartz-etal-2020-unsupervised}.
Differing from \citet{shwartz-etal-2020-unsupervised} heuristic designs for sample patterns of different datasets, our $\mathcal{T}_K$ is based on our hierarchical comprehension skills and collaborates with the question focus to develop a complete chain of thought of PLM.
To be specific, the prompt input $\mathcal{P}_K(c, f)$ in Eq.~\ref{eq:knowledge-generation} of skill-specific knowledge generation is represented as:
\begin{equation*}
    \mathcal{P}_K (c, f) = \inlinetcbox[boxsep=5pt]{c} \text{\texttt{From the context:}} \inlinetcboxmathcalandothers{T}{F}{(f)} \inlinetcboxmathcal{T}{K}
\end{equation*}
where $\mathcal{T}_{F}(f)$ means infilling the $\mathcal{T}_{F}$ with corresponding generated question focus $f$.


\noindent
\textbf{Model training.} \
To augment the original input context, we first fill the \textit{F-template} and \textit{K-template} with the generated question focus and knowledge text.
After that, we append them to the original context to obtain the augmented input:
\begin{equation*}
    \text{Aug}(c, f, k) = \inlinetcbox[boxsep=5pt]{c} \inlinetcboxmathcalandothers{T}{F}{(f)} \inlinetcboxmathcalandothers{T}{K}{(k)}
\end{equation*}
Furthermore, to help our \texttt{SkillQG} learn the relationship between multiple pieces of input text and capture their functions, we utilize natural language prompts as well as special tokens as the delimiter to combine the multiple inputs into a single sequence, \ie including the knowledge-augmented context $c$, answer text $a$, and skill $s$.
This kind of method has been proven to help better learn the relationship between multiple pieces of input text and capture their functions, improving performance on various tasks~\cite{schick-schutze-2021-exploiting,zhou-etal-2022-think}.
Formally, the input sequence fed into our question generator is as follows:
\begin{equation*}
\text{\texttt{[CXT]}} \inlinetcbox[boxsep=5pt]{c} \text{\texttt{[ANS]}} \inlinetcbox[boxsep=5pt]{a} \text{\texttt{[SKL]}} \inlinetcbox[boxsep=5pt]{s} \text{\texttt{Ask a question:}} 
\end{equation*}
where \texttt{[CXT]}, \texttt{[ANS]} and \texttt{[SKL]} are special tokens to mark the boundary between multiple input sequences~\cite{radford2019language}.
After that, the sequence is fed into a BART-base~\cite{lewis-etal-2020-bart} question generator which models the probabilities $p_\theta \left(q \mid c, s, a \right)$ in Eq.~\ref{eq:generation} by minimizing the conditional negative log-likelihood (NLL) loss:
\begin{align}
    \mathcal{L}_{QG} = -\sum_{t=1}^{T}{\log \hat{p_\theta} \left(q_t \mid q_{<t}, c, s, a \right)}
\end{align}
where $\hat{p_\theta} \left(q_t \mid q_{<t}, c, s, a \right)$ denotes the predicted probability for the token in the reference question.

\section{Experiments}
\label{sec:experiments}

\noindent
\textbf{Datasets.} \
We employ the official train and dev splits of FairytaleQA dataset~\cite{xu-etal-2022-fantastic} to train our \texttt{SkillQG}.
This dataset, focusing on narrative comprehension of English text for both machines and young children, is annotated with seven fine-grained skills comprised of Character, Setting, Action, Feeling, Causal relationship, Outcome resolution, and Prediction.
Its annotation process is supervised by three experts in literacy education and its categorization of questions is based on prior educational research~\cite{paris2003assessing} so that we can easily match the samples of the FairytaleQA dataset with our defined skill schema.
Table~\ref{tab:fairytaleqa-bloom} presents this mapping relationship and corresponding breakdown statistics of the dataset.

\begin{table}[!t]
    \centering
    \resizebox{1.0\columnwidth}{!}{
        \begin{tabular}{cccc}
            \toprule
            \textbf{Annotation}& \textbf{Count}& \textbf{Percentage (\%)}& \textbf{Skill} \\
            \midrule
            \midrule
            Character& 1172& 11.08& \textsc{Remember} \\
            Setting& 630& 5.95& \textsc{Remember} \\
            Action& 3342& 31.59& \textsc{Understand} \\
            Feeling& 1024& 9.68& \textsc{Evaluate} \\
            Causal rel.& 2940& 27.79& \textsc{Analyze} \\
            Outcome res.& 986& 9.42& \textsc{Analyze} \\
            Prediction& 486& 4.59& \textsc{Create} \\
            \bottomrule
        \end{tabular}
    }
    \caption{Breakdown statistics of the FairytaleQA dataset and its mapping to our proposed skill-based schema.}
    \label{tab:fairytaleqa-bloom}
    \vspace{-5mm}
\end{table}

\noindent
\textbf{Baselines.} \
We compare SkillQG to the following two types of QG baselines.
The first type is typically trained without the knowledge input, including NQG++~\cite{zhou2017neural}, and QAG~\cite{yao-etal-2022-ais}.
The other is knowledge-augmented generators consisting of CsQG~\cite{xin-etal-2021-enhancing} and CQG~\cite{fei-etal-2022-cqg}, which retrieve external knowledge from knowledge bases or generate knowledge with another model and regard the knowledge as extra context to generate questions.

\subsection{Evaluation Protocol}
\label{subsec:evalution-prococol}

\noindent
\textbf{Automatic evaluation metrics.} \
We use standard question generation metrics to evaluate the question quality from the following three aspects.
The \textbf{syntactic similarity} between generated questions and reference is measured by BLEU-4~\cite{papineni-etal-2002-bleu} and ROUGE-L~\cite{lin-2004-rouge}.
The \textbf{answerability} and structural integrity of generated questions is gauged by Q-BLEU-4~\cite{nema-khapra-2018-towards}.
The \textbf{relvance} of generated questions to the 
reference is evaluated by BERTScore~\cite{zhang2019bertscore}, while that to the given context is evaluated by the factuality dimension of CTC~\cite{deng-etal-2021-compression} and BARTScore~\cite{yuan2021bartscore}.

\noindent
\textbf{Human evaluation.} \
We conduct a voluntary human evaluation to analyze \texttt{SkillQG} by asking five annotators to rate the quality of candidates generated by different models when using 300 $\langle passage, skill, answer, question \rangle$ samples in the unseen test split as the input.
For \textbf{question content quality}, following the human criteria of QG elaborated by \citet{rus-etal-2010-first} and \citet{nema-khapra-2018-towards}, we conduct pairwise comparison 
where we present a context and two questions made by two different models and ask the annotators to choose the better of the two or ``tie'' in terms of grammaticality, answerability, and relevance.
We report the percentage of times annotators prefer each model to NQG++ and ties, \ie wins/ties ratio.
For \textbf{skill controllability}, we ask the annotator to read the context, the generated questions, and the corresponding answer, choose the evidence sentences in context, and then respectively annotate the required comprehension skill from our defined 5-dimensional skill schema.
\emph{Please refer to Appendix~\ref{sec:annotation-details} for more details about the annotation.}


\begin{table}[!t]
    \centering
     \resizebox{1.0\columnwidth}{!}{
        \begin{tabular}{c|c|cc|ccc}
            \toprule
            \multicolumn{1}{c}{\textbf{Method}}& 
            \centeroneg{\textbf{Q-B4}}& \centeroneb{\textbf{R-L}}& \centeroneb{\textbf{B4}}& \centeroner{\textbf{BE.S}}& \centeroner{\textbf{CTC}}& \centeroner{\textbf{BA.S}} \\
            \midrule
            \midrule
                NQG++& 0.503& 0.421& 0.141& 0.342& 0.328& 0.266 \\
                QAG& 0.552& 0.427& 0.146& 0.424& 0.408& 0.333 \\
                QTD& 0.576& 0.431& 0.150& 0.478& 0.456& 0.372 \\
                \hline
                CsQG& 0.592& 0.431& 0.151& 0.506& 0.485& 0.393 \\
                CQG& 0.609& 0.433& 0.153& 0.532& 0.510& 0.415 \\
                \hline
                \textbf{SkillQG}& \textbf{0.656}& \textbf{0.440}& \textbf{0.159}& \textbf{0.620}& \textbf{0.596}& \textbf{0.485} \\
            \bottomrule
        \end{tabular}
     }
    \caption{Quantitative results in terms of answerability, syntactic similarity, and relevance evaluation metrics on the FairytaleQA dataset.
    Please refer to Section~\ref{subsec:evalution-prococol} for the full name of employed metrics. 
    The best result is marked as \textbf{bold}.
    }
    \label{tab:main-results}
    \vspace{-2mm}
\end{table}

\subsection{Main Results}
\label{subsec:main-results}
Table~\ref{tab:main-results} summarizes the quantitative results on the FairytaleQA dataset.
On the one hand, compared with the baselines without extra knowledge (\ie NQG++, QAG, QTD), \texttt{SkillQG} achieves obviously higher metrics scores in terms of answerability, and relevance, demonstrating the significant contribution of incorporating extra knowledge and question focuses to generate the questions.
The comparable results on syntactic similarity metrics may be attributed to the wrong penalization of these metrics to the novel generation of our \texttt{SkillQG}.
On the other hand, \texttt{SkillQG} consistently outperforms all the knowledge-augmented baselines (\ie CsQG and CQG) by a considerable margin (\ie gain ratio $> 5\%$), which indicates the effectiveness of externalized knowledge by our devised prompts.


\begin{table}[!t]
    \centering
     \resizebox{1.0\columnwidth}{!}{
        \begin{tabular}{c|cc|cc|cc}
            \toprule
            \multirow{2}{*}{Method}& \multicolumn{2}{c|}{\textbf{Grammaticality}}& \multicolumn{2}{c|}{\textbf{Answerability}}& \multicolumn{2}{c}{\textbf{Relevance}} \\
            & \textbf{wins\%}& \textbf{ties\%}& \textbf{wins\%}& \textbf{ties\%}& \textbf{wins\%}& \textbf{ties\%} \\
            \midrule
            \midrule
            QAG& 46.3& 8.7& 47.0& 9.0& 41.7& 20.5 \\
            QTD& 48.7& 9.3& 48.3& 2.7& 48.2& 6.3 \\
            \hline
            CsQG& 49.0& 7.3& 49.2& 5.3& 49.4& 6.0 \\
            CQG& 50.3& 4.0& 51.3& 5.0& 52.0& 7.0 \\
            \textbf{SkillQG}& \textbf{53.0}& 10.0& \textbf{53.6}& 5.6& \textbf{54.0}& 3.7 \\
            \bottomrule
        \end{tabular}
     }
    \caption{Human evaluation results on question content quality.
    We show the percentage of times annotators prefer each variant to NQG++ and ties.}
    \label{tab:question-quality}
    \vspace{-5mm}
\end{table}

\noindent
\textbf{Inter-annotator agreement.} \
For the examined two aspects of human evaluation, \ie question content quality and skill-controllability, the inter-annotator Krippendorff's $\alpha$ for them are 87.20 and 90.73, respectively, which demonstrates an acceptable level of agreement (> 80\%) between annotators~\cite{krippendorff2004reliability}.
The annotators discuss the few annotation conflicts to reach a unanimous conclusion.

\noindent
\textbf{Question content quality.} \
As shown in Table~\ref{tab:question-quality}, the pairwise comparisons show that \texttt{SkillQG} produces more grammatical and relevant questions and questions that are mostly answerable ($> 50\%$), compared to all baseline models.
Besides, knowledge-augmented baselines (lower part in Table~\ref{tab:question-quality}) consistently receive more preference from annotators than others (upper part in Table~\ref{tab:question-quality}).
It demonstrates that the generated skill-specific knowledge indeed enhances the question content and relevance.

\begin{figure}[!t]
\centering
    \includegraphics[width=0.8\columnwidth]{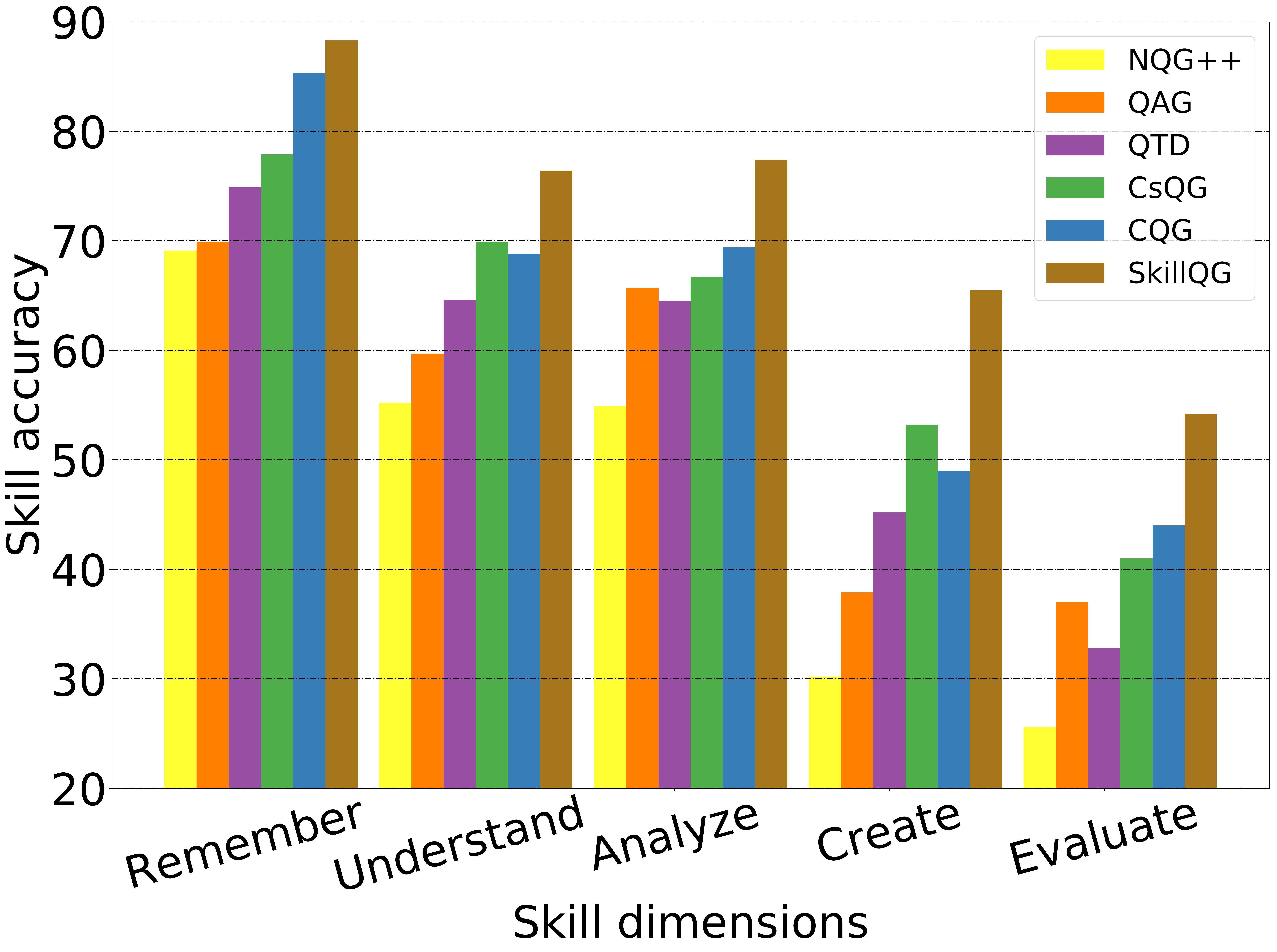}
    \caption{Human evaluation results on skill controllability, which is computed by comparing the given with the annotated skill.
    We depict the accuracy for each skill alongside the horizontal axis.
    }
    \label{fig:skill-controllability}
    \vspace{-5mm}
\end{figure}

\noindent
\textbf{Skill controllability.} \
Figure~\ref{fig:skill-controllability} reports the consistency between the given skill name that \texttt{SkillQG} generates questions conditioned on and the one chosen by the annotators, \ie skill accuracy.
We can see our \texttt{SkillQG} surpasses other baselines by a significant margin, and this becomes more obvious to the skills that have a relatively smaller number of samples in the dataset, \ie around 30\% gain in \textsc{Create} and \textsc{Evaluate} dimension.
It justifies that \texttt{SkillQG} can not only successfully control the comprehension skill of generated questions, but also be able to learn the underrepresented skills in the dataset, owing to the built prompts containing indicative words of different comprehension skills and the rich skill-specific knowledge of language models.
\emph{Please refer to Section~\ref{sec:more-experiments} for more results.}


        
            
        
        
        
        
        


        
        


\subsection{Ablation Analysis}

We conduct ablation experiments and summarize the results in Table~\ref{tab:ablation-results} from the following aspects.

First, \textit{How do the special symbols and prompts of input representation contribute to the generation quality?}
The first three baselines combine multiple input sequences (\ie context, answer, and skill) with the concatenation operation, special symbols or natural language prompts, denoted as ``concat-only ($M_1$)'', ``symbol-only ($M_2$)'' and ``prompt-only ($M_3$)'', respectively.
As shown in Table~\ref{tab:ablation-results}, we can observe that $M_1$ achieves worse performance than $M_2$ and $M_3$, demonstrating that simple concatenation operation cannot encode the input sequences well.
Besides, both $M_2$ and $M_3$ degrade the performance w.r.t. \texttt{SkillQG}, showing the integration of special symbols and natural language prompts can help the generator better understand the relationship between multiple input sequences and improve the final quality.

\begin{table}[!t]
    \centering
     \resizebox{1.0\columnwidth}{!}{
        \begin{tabular}{c|c|cc|ccc}
            \toprule
            \multicolumn{1}{c}{\textbf{Baseline}}&
            \centeroneg{\textbf{Q-B4}}& \centeroneb{\textbf{R-L}}& \centeroneb{\textbf{B4}}& \centeroner{\textbf{BE.S}}& \centeroner{\textbf{CTC}}& \centeroner{\textbf{BA.S}} \\
            \midrule
            \midrule
            concat-only ($M_1$)& 0.626& 0.436& 0.155& 0.569& 0.550& 0.445 \\
            symbol-only ($M_2$)& 0.639& 0.437& 0.156& 0.581& 0.562& 0.457 \\
            prompt-only ($M_3$)& 0.641& 0.438& 0.158& 0.598& 0.572& 0.466 \\
            \hline
            generator ($M_4$)& 0.620& 0.434& 0.155& 0.558& 0.536& 0.435 \\
            conceptnet ($M_5$)& 0.636& 0.436& 0.156& 0.582& 0.559& 0.455 \\
            \hline
            \textbf{SkillQG}& \textbf{0.656}& \textbf{0.440}& \textbf{0.159}& \textbf{0.620}& \textbf{0.596}& \textbf{0.485} \\
            \bottomrule
        \end{tabular}
     }
    \caption{Quantitative results of ablation experiments.
    The best result is marked as \textbf{bold}.
    }
    \label{tab:ablation-results}
    \vspace{-5mm}
\end{table}

Second, \textit{What is the impact of question focus and skill-specific knowledge?}
The baseline ``generator ($M_4$)'' does not utilize skill-specific knowledge to augment the context and trains the question generator directly, \ie a BART model for question generation, while the baseline ``conceptnet ($M_5$)'' is trained in the similar setting to \texttt{SkillQG} but its extra knowledge is attained by retrieving the ConceptNet rather than inquiring PLM.
We perform alignment between the context and ConceptNet following the embedding-based matching as \citet{zhou-etal-2022-think}.
In Table~\ref{tab:ablation-results}, we can find that the contribution of extra knowledge from PLM (\texttt{SkillQG} v.s. $M_4$) is more significant than that from the ConceptNet ($M_5$ v.s. $M_4$).
A possible reason is that chain-of-thought prompting of PLM can reflect better relevance and specificity of knowledge to the given context and the required comprehension skill compared to matching with the limited number of triplets in a knowledge base.
This result also agrees with the recent study on evaluating PLM as a knowledge base~\cite {heinzerling-inui-2021-language}.


\subsection{Boosting QA Performance using Unlabeled Corpus}
\label{subsec:augmented-qa}

We further evaluate whether the skill-controllable questions can improve QA performance through data augmentation and help us better understand the QA models' intrinsic ability.
Specifically, we first devise an information extractor to obtain $\langle passage, skill, answer \rangle$ combinations on an unlabeled corpus, \ie the passage without annotations of the question, answer, and skill.
After that, we feed the extracted combinations of $\langle$ \emph{passage}, \emph{answer}, \emph{skill} $\rangle$ into \texttt{SkillQG} to generate skill-infused questions.
Finally, we put the generated questions into the FairytaleQA training set and train a QA model with such an augmented dataset to further evaluate the effectiveness of our \texttt{SkillQG}.

\noindent
\textbf{Information extraction.}
Since the answer and required skill are dependent on each other, we cannot sample the combinations of $\langle$ \emph{passage}, \emph{answer}, \emph{skill} $\rangle$ randomly.
Following the widely adopted solutions~\cite{liu2020asking, ghanem-etal-2022-question}, we decompose the process into two steps to sequentially sample the required skills, and corresponding answers to select reasonable combinations.
Formally, the sampling procedure can be written as:
\begin{align}
    p \left(s, a \mid c \right) = p \left(s \mid c \right) p \left( a \mid c, s \right)
\end{align}
where $p \left(s \mid c \right)$ and $p \left( a \mid c, s \right)$ are devised as a model-based and rule-based extractor, respectively.

On the one hand, $p \left(s \mid c \right)$ is formulated as a multi-label classification task because a passage may involve more than one skill.
We first fine-tune a DistilBERT model~\cite{sanh2019distilbert} on the FairytaleQA dataset to learn skill-related patterns in the context.
After that, we use it to predict the candidate skills when given an unlabeled passage.

On the other hand, we borrow the statistical analysis on the FairytaleQA dataset from \citet{yao-etal-2022-ais} and implement $p \left( a \mid c, s \right)$ using heuristic rules.
Specifically, \textsc{Remember} and \textsc{Evaluate} skills, \ie the narrative elements consisting of character, setting, and feelings, are usually based on the named entities, such as a mentioned name and a particular place.
Therefore, we resort to the Spacy tool~\cite{honnibal2017natural} to extract named entities as the candidate answers.
Other skills, \ie the narrative elements consisting of action, causal relationship, outcome resolution, and prediction are mainly made up of the action events.
Thus, we first leverage Propbank's semantic role labeler~\cite{johansson-nugues-2008-dependency} to extract the trigger verb as well as the involved subject and object and then concatenate them into a complete sentence as the candidate answers.

We conduct the sampling procedure on the passages of FairytaleQA training set and discard all their annotations, then feed the extracted $\langle$ \emph{passage}, \emph{answer}, \emph{skill} $\rangle$ into \texttt{SkillQG} by keeping all beam search
(size=8) outputs for each sample.
Consequently, we can generate diverse questions for the existing paragraphs in the FairytaleQA training set.
Finally, we randomly select 80,000 candidate questions and augment the FairytaleQA training set with them.
As a comparison, following the same setting as above, we design a baseline by utilizing CQG as the question generator, which is one of the most competitive metrics in Table~\ref{tab:main-results}.

\begin{figure}[!t]
    \centering
	\begin{subfigure}{0.49\columnwidth}
        \includegraphics[width=\columnwidth]{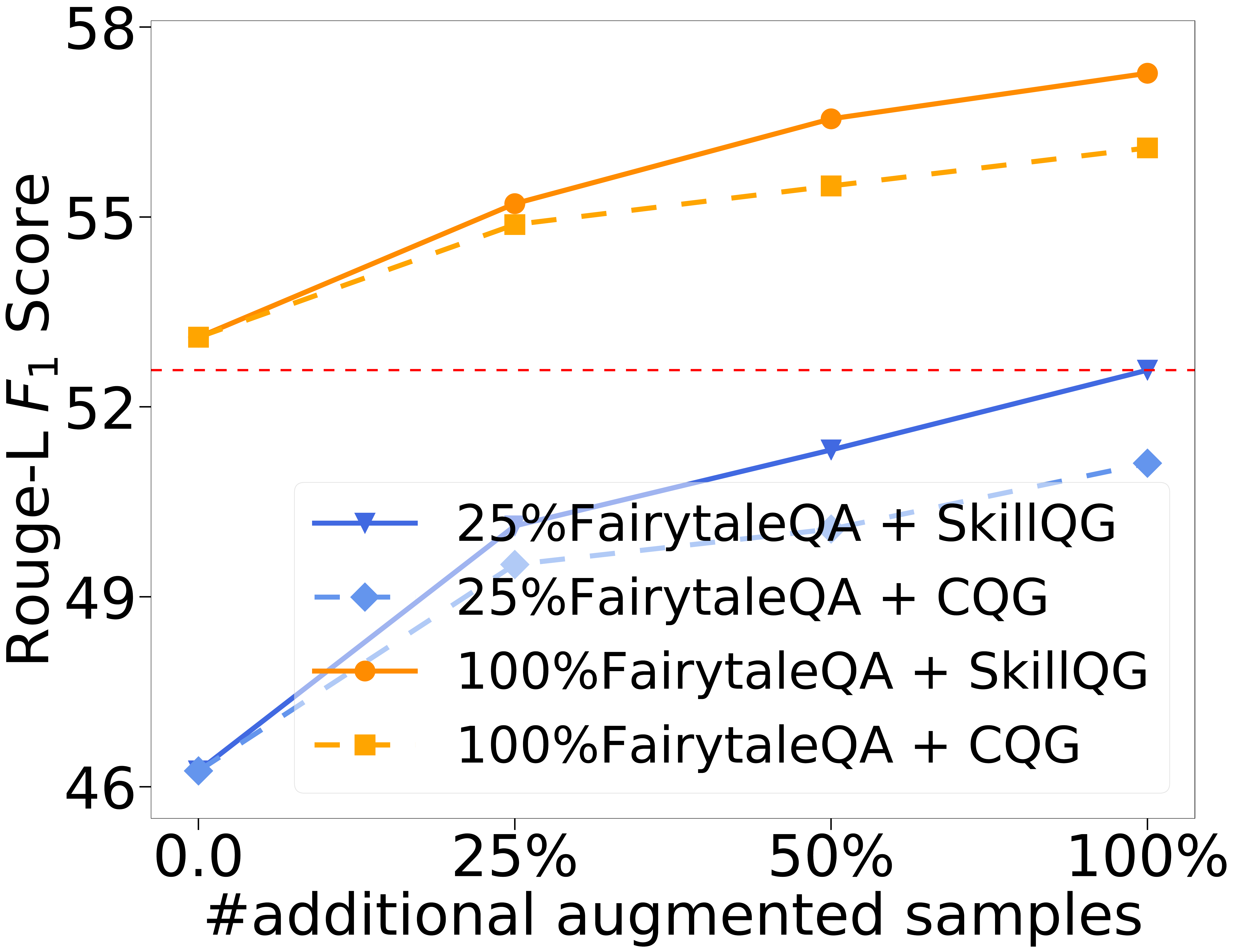}
        \subcaption{Comparison of different question generators.}
        \label{fig:augmented-qa}
	\end{subfigure}
	\begin{subfigure}{0.49\columnwidth}
        \includegraphics[width=\columnwidth]{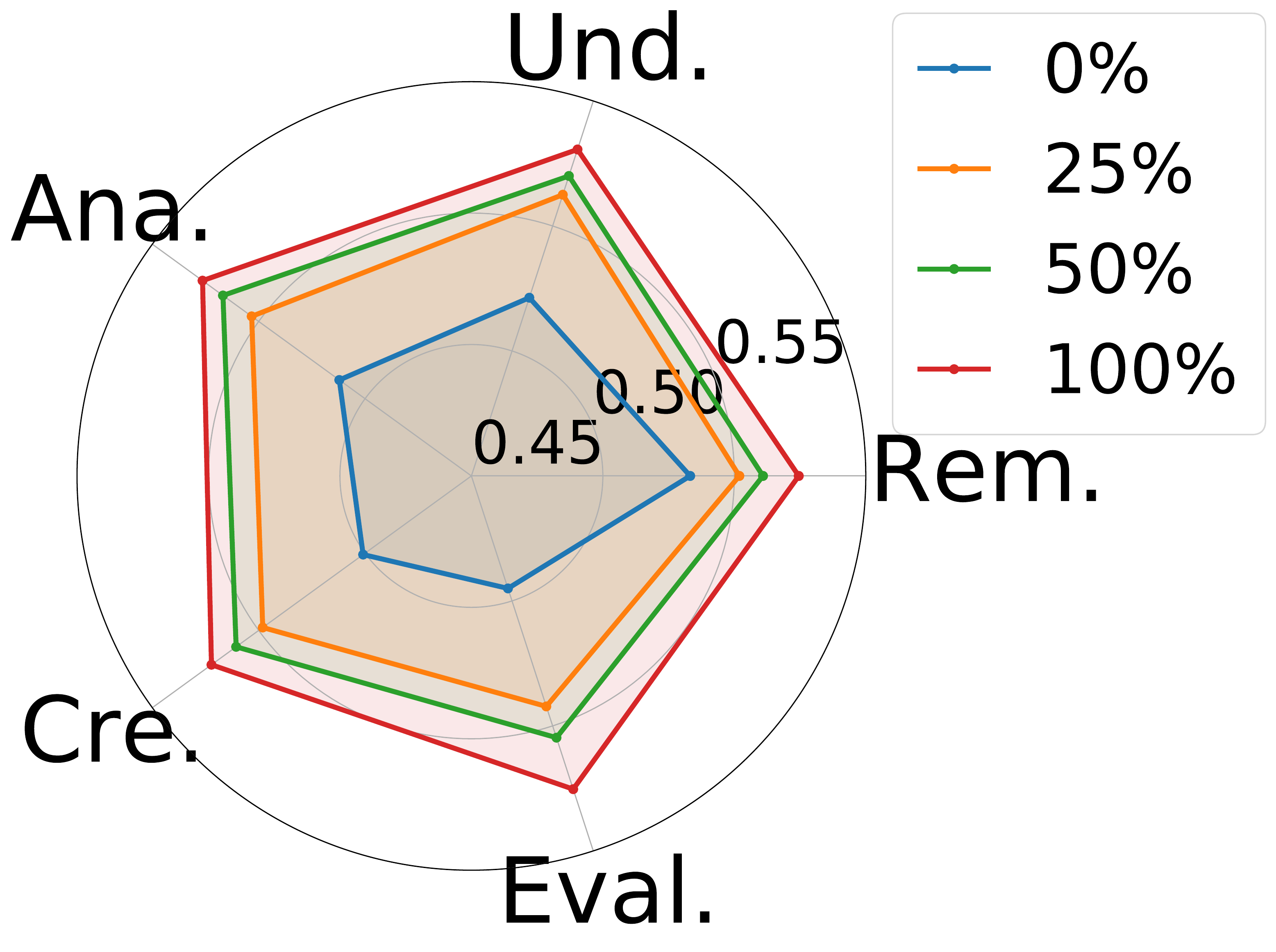}
        \subcaption{Breakdown analysis of the overall QA performance.}
        \label{fig:augmented-qa-decomposition}
    \end{subfigure}
    \caption{Overall and decomposed performance of the state-of-the-art QA model on the FairytaleQA dataset, augmented with data generated by question generators.
    }
    \vspace{-5mm}
\end{figure}

We train a state-of-the-art QA baseline~\cite{xu-etal-2022-fantastic} on the augmented dataset to further evaluate the quality of generated questions.
Following \citet{xu-etal-2022-fantastic}, we report the QA performance in Rouge-L $F_1$ Score which is a commonly used metric for generative question answering.
The results in a high-resource setting (with the whole FairytaleQA training set) and a low-resource setting (with only 25\% of data sampled from the original FairytaleQA training set) are illustrated in Figure~\ref{fig:augmented-qa}.
We can observe that the questions generated by \texttt{SkillQG} can improve the QA performance to a greater extent than CQG under both settings.
In particular, the QA model under the low-resource setting achieves a comparable performance to the high-resource setting when leveraging the 100\% additional samples generated by our \texttt{SkillQG}.

Furthermore, Figure~\ref{fig:augmented-qa-decomposition} illustrates the decomposed performance of \texttt{SkillQG} under the low-resource setting (\ie ``25\%FairytaleQA + SkillQG'' setting shown in Figure~\ref{fig:augmented-qa}
) alongside the defined skill dimension.
This result shows that the questions generated by \texttt{SkillQG} can significantly boost all of the comprehension capabilities for the QA model.
Among them, the cognitively challenging ones that the QA model struggles in, such as \textsc{Evaluation} and \textsc{Create}, even achieve the largest improvement.
It demonstrates that the skill-controllable questions that generated by the \texttt{SkillQG} can compensate for the limited number of training samples in the FariytaleQA dataset and are favorable for the fine-grained assessment of comprehension capability of QA models.

\section{Related Work}
\label{sec:related-work}
\noindent
\textbf{Deep question generation.} \
Previous QG systems mainly generated factoid-based questions by a sequence-to-sequence model~\cite{zhou2017neural, liu2019learning}, a PLM~\cite{liu2020asking}, or a graph-based architecture~\cite{talmor-berant-2018-web, kumar2019difficulty}.
Recent-emerged QG models aimed at generating questions that require deep reasoning.
On the one hand, \citet{cheng-etal-2021-guiding} proposed to generate difficulty-controllability questions through step-by-step rewriting, while \citet{bi-etal-2021-simple-complex} decoded multi-hop questions by a soft template.
On the other hand, \citet{yao-etal-2022-ais} and \citet{zhao-etal-2022-educational} devised educational question generators to facilitate the assessment of children's literacy.
Our \texttt{SkillQG} is inspired by their fine-grained analysis but driven by the motivation that generating questions with deep comprehension is beneficial to QA training.
More recently, \citet{cao-wang-2021-controllable} charted a new question ontology, but they focused on constructing diversified open-ended questions from the specified question types.

\noindent
\textbf{Knowledge-augmented generation.} \
Although explicit knowledge generation has been explored in natural language understanding~\cite{liu-etal-2022-generated, wei2022chain}, similar research on natural language generation~\cite{zhou-etal-2022-think}, especially for QG is relatively rare~\cite{rajani-etal-2019-explain}.
\citet{xin-etal-2021-enhancing} retrieved knowledge triplets from ConceptNet~\cite{speer2017conceptnet} to enhance the QG models, while \citet{fei-etal-2022-cqg} adopted a graph attention networks (GAT)~\cite{velivckovic2018graph} to capture focuses for question generators.
We considered the lessons of these works and extend the knowledge source with pre-trained language models.

\section{Conclusion}
\label{sec:conclusion}
Existing QG systems focus on the literal nature of questions and rarely consider the comprehension types of the generated questions.
To better assess and improve machine reading comprehension models, we propose \texttt{SkillQG} to generate questions with controllable comprehension types.
Besides, we engage the question focus and specific knowledge to improve the controllability of generation.
Empirical results show that \texttt{SkillQG} outperforms baselines while achieving a significant performance boost in downstream QA training.

\section{Limitations}
Our work proposes a new QG framework, namely \texttt{SkillQG}, to frame the comprehension skill required by a question and generate the corresponding comprehension-oriented questions. The limitations are three-fold:

Firstly, we propose a new skill-based schema for the comprehension nature of questions and map the existing annotations on narrative elements of the FairytaleQA dataset to it and conduct our experiments.
This kind of mapping might not reflect the required skills accurately since a narrative element can cover more than one comprehension types.
One remedy to this issue could be collecting a new QA dataset with the annotations following our proposed schema.
We regard it as our future work and deem designing a new annotation specification a promising direction.

Besides, although we boost the downstream QA performance in Section~\ref{subsec:augmented-qa} by augmenting the original training set with generated questions, the final performance (56.9\%) is also far behind the human performance (64.4\%) reported by \citet{xu-etal-2022-fantastic}.
However, the breakdown analysis of QA performance demonstrates that \texttt{SkillQG} can strengthen all of the comprehension capabilities, especially the challenging ones.
As a result, generating questions that are matched with the current comprehension capabilities of the QA model and co-evolving the QA system and corresponding QG system, could be two interesting research topics. 

Last but not least, our \texttt{SkillQG} is built on the PLMs of general domains, ignoring the domain-specific and multilingual application.
The backbone PLMs are also shown a biased representation, such as race and gender~\cite{gonen-goldberg-2019-lipstick-pig}.
Therefore, additional evaluation protocols are left for our future work. 

\section*{Acknowledgements}
We would like to thank anonymous reviewers for their valuable comments and suggestions.
This work has been supported in part by the NSFC (No. 62272411), the Zhejiang NSF (LR21F020004), Ant Group and Alibaba-Zhejiang University Joint Research Institute of Frontier Technologies.

\bibliography{anthology,custom}
\bibliographystyle{acl_natbib}

\clearpage

\appendix

\section{Focus and Knowledge Templates}
\label{sec:fk-templates}

\begin{table*}[!t]
    \centering
    \resizebox{1.0\textwidth}{!}{
        \begin{tabular}{cll}
            \toprule
            \textbf{Skill}& \multicolumn{1}{c}{\textbf{\textit{F-template} \pmb{$\mathcal{T}_{F}$}}}& \multicolumn{1}{c}{\textbf{\textit{K-template} \pmb{$\mathcal{T}_{K}$}}} \\
            \midrule
            \midrule
            \multirow{3}{*}{\textsc{Remember}}& What is the definition of <blank>& The definition of <focus> is <blank> \\
            & What are the properties of <blank>& The properties of <focus> are <blank> \\
            & How would you describe <blank>& <focus> is a <blank> \\
            \hline
            \multirow{6}{*}{\textsc{Understand}}& What is the purpose of <blank>& The purpose of <focus> is to <blank> \\
            & What is the main function of <blank>& The main function of <focus> is <blank> \\
            & How would you classify the type of <blank>& The type of <focus> is <blank> \\
            & What is the difference between <blank>& The difference between <focus> is <blank> \\
            & How would you rephrase the meaning of <blank>& The meaning of <focus> is <blank>\\
            & How would you summarize <blank>& The summarization of <focus> is <blank>\\
            \hline
            \multirow{4}{*}{\textsc{Analyze}}& How would <blank> feel afterwards?& <focus> felt <blank> \\
            & What happened as a result of <blank>& As a result of <focus>, <blank>\\
            & What might have caused <blank>& The cause of <focus> was <blank> \\
            & Why did <blank> do this?& <focus> did this because they wanted <blank> \\
            \hline
            \multirow{4}{*}{\textsc{Create}}& What will <blank> want to do next?& <focus> want <blank> \\
            & What will happen to <blank> next?& <focus> will <blank> \\
            & What would happen if <blank>& If <focus>, <blank> \\
            & What will be the outcome if <blank>& If <focus>, the output will be <blank> \\
            \hline
            \multirow{2}{*}{\textsc{Evaluate}}& Why do you recommend <blank>& You recommend <focus> because <blank> \\
            & Why is it better that <blank>& It is better that <focus> because <blank> \\
            \bottomrule
        \end{tabular}
    }
    \caption{\textit{F-template} and \textit{K-template} used for each defined comprehension skills.}
    \label{tab:fk-templates}
    \vspace{-5mm}
\end{table*}

We manually design  a few generic templates to conduct chain-of-thought prompting to the PLM, which are with a form of information-seeking questioning pairs.
Table~\ref{tab:fk-templates} summarizes the employed focus and knowledge templates, \ie \textit{F-template} and \textit{K-template}, where ``<blank>'' means the placeholder to be filled with the generated question focus and knowledge text, while ``<focus>'' represents the question focus text.
In addition to the prefix-style templates, \ie ``<blank>'' is located in the trailing part of the input prompt, we also resort to cloze-style templates, \ie ``<blank>'' is in the middle part of the input prompt, such as ``\textit{How would <blank> feel afterwards}?''.
Our employed PLM, \ie GPT2, is based on causal language modeling, and does not well in finishing such cloze-style template.
Therefore, we leverage Spacy tool~\cite{honnibal2017natural} to extract named entities as the generated question focus for these prompts and follow the same pipeline elaborated in Section~\ref{subsec:skill-qg}.

As show in the table, each comprehension skill is equipped with at least 2 \textit{F-template}s.
We generate 5 question focuses for each \textit{F-template} using Nucleus sampling~\cite{holtzman2019curious} with well-adopted p = 0.2, \ie sampling from the top 20\% tokens~\cite{holtzman2019curious} and obtain the full question focus when the eos token generates.
In addtion, each \textit{F-template} is paired with a corresponding \textit{K-template}.
We use Nucleus sampling with p = 0.5 to generate 10 pieces of knowledge text for each \textit{K-template}.

\section{Implementations Details}
\label{sec:implementation-details}
The question generator of our \texttt{SkillQG} is built on the basis of a BART-base model~\cite{lewis-etal-2020-bart}, while the skill-specific knowledge is generated by a GPT2 model~\cite{radford2019language}.
The number of their parameters are around 140M and 117M, respectively. 
Both of them are frist initialized by the pre-trained parameters of the HuggingFace Transformers package~\cite{wolf-etal-2020-transformers}.
After that, the parameters of the GPT2 model will be frozen and ones of the BART-base model will be fine-tuned on the training set.
Our information extractor is initialized by a DistilBERT model with about 66M parameters from the HuggingFace package.
AdamW~\cite{loshchilov2018decoupled} optimizer with weight decay 5e-4 and epsilon 8 is used to fine-tune the model with a maximum sequence length of 384.
During training, we extract mentioned sections of a whole passage as the input context, which is annotated in the $corr\_sec$ field of the FairytaleQA dataset.
The learning rate warms up over the first 10\% steps and then decays linearly to 0 for all experiments with training batch size 16 and maximum iteration 40,000.
The whole training takes 25 hours on 4 NVIDIA GTX 2080Ti GPUs.
We use the official train split to fine-tuning the question generator of our \texttt{SkillQG} and employ grid search to determine the hyper-parameters based on the val split.
We report the average results of ten runs for automatic evaluation and conduct the human evaluation on the candidates generated by a single run.

\begin{table*}[!t]
    \centering
     \resizebox{1.0\textwidth}{!}{
        \begin{tabular}{c|l|c|l}
            \toprule
            \multicolumn{2}{c}{\textbf{Candidate question}}& \multicolumn{1}{c}{\textbf{Instruction}}& \multicolumn{1}{c}{\textbf{Description}} \\
            \hline
            \multirow{2}{*}{\textbf{Grammaticality}}& \textbf{A}. How many solo tackles did Von Miller make at Super Bowl?& \multirow{2}{*}{A wins B.}& \multirow{2}{*}{B is not grammatically correct.} \\
            & \textbf{B}. What site is \textcolor{dred}{locate} in the San Franc?& & \\
            \hline
            \multirow{3}{*}{\textbf{Answerability}}& \textbf{A}. How many Grammys has Lady Gaga won?& \multirow{3}{*}{A wins B.}& \multirow{3}{8cm}{B misses some important information, such as named entities, relation words, and question words.} \\
            & \multirow{2}{9cm}{\textbf{B}. How many \textcolor{dred}{\st{professors}} does the Warsaw University of Technology employ?}& & \\
            & & & \\
            \hline
            \multirow{6}{*}{\textbf{Relevance}}& \textbf{A}. What is the axis of Warsaw which divides it into two parts?& \multirow{6}{*}{A wins B.}& \multirow{6}{8cm}{B is partially relevant but unable to be grounded by the context.} \\
            & \multirow{2}{9cm}{\textbf{Context of A}. $[\ldots]$ the Vistula River is the specific axis of Warsaw, which divides the city into two parts $[\ldots]$}& & \\
            & & & \\
            & \textbf{B}. \textcolor{dred}{How big} is the greater metropolitan area?& & \\
            & \multirow{2}{9cm}{\textbf{Context of B}. $[\ldots]$ within a greater metropolitan area of 2.666 million residents $[\ldots]$}& & \\
            & & & \\
            \bottomrule
        \end{tabular}
     }
    \caption{Scoring examples for the human evaluation on the question content quality.
    The problematic words in corresponding candidate questions are marked in \textcolor{dred}{red}.
    }
    \label{tab:question-quality-annotation-examples}
\end{table*}

\section{Annotation Details}
\label{sec:annotation-details}
\begin{figure*}[!t]
\centering
    \includegraphics[width=0.9\textwidth]{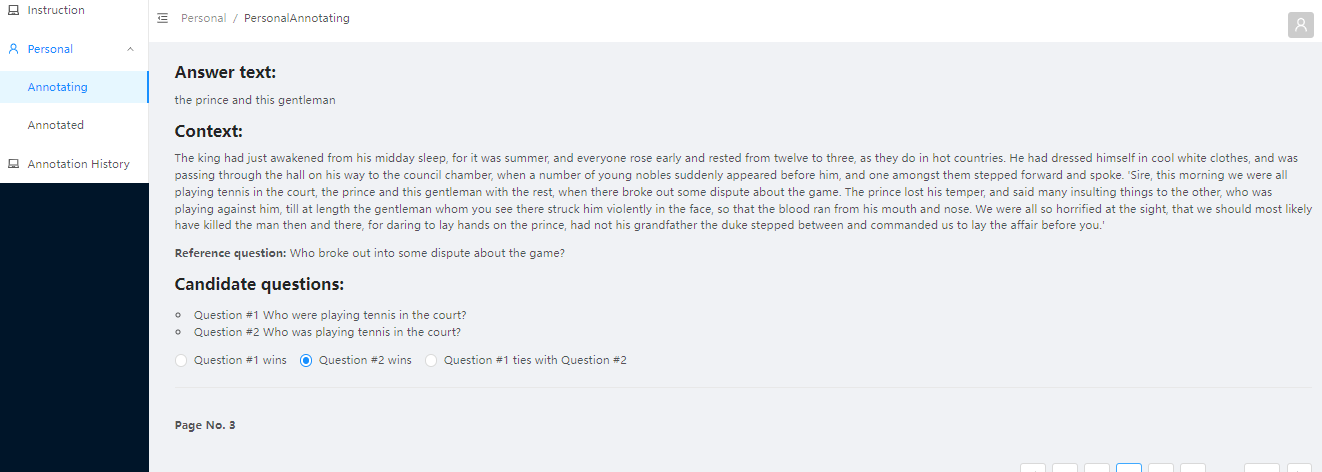}
    \caption{A screenshot of our human annotation process.}
    \label{fig:software}
\end{figure*}

Our human evaluation is conducted by a total of five annotators.
All of the annotators are from China, between 25 and 30 years old, competent in English and studying as Computer Science graduates.
They are informed of the necessary background knowledge on QG and evaluation for QG, as well as detailed annotation instructions along with examples when participating our study.
In addition, they gladly volunteered to provide their assistance without being compensated in any form.
The candidate questions are anonymized and evaluated in the following aspects:
\begin{itemize}
    \item \textbf{Question content quality.} Following the human criteria elaborated in QG-STEC Task B~\cite{rus-etal-2010-first}, we check whether a question is well-formed, answerable, and relevant to the context.
    Besides, previous works have shown that pairwise comparison produces a more reliable evaluation than directly asking humans to score the candidate~\cite{amidei2019use, celikyilmaz2020evaluation}.
    Therefore, we present a context and two questions made by two different models and ask the annotators to choose the better of the two or ``tie''.
    Specifically, we first show the annotators a candidate question generated by NQG++ and a one generated by others as well as the corresponding input context and answer text.
    After that, we ask the annotators to compare the two questions in terms of grammaticality, answerability, and relevance.
    To better guide the annotators to distinguish between high-quality candidate questions and low-quality ones, we also show the annotators clearing examples as presented in Table~\ref{tab:question-quality-annotation-examples}.
    
    \item \textbf{Skill-controllability.} It checks the consistency between the given skill that question generator are conditioned on and the one chosen by the annotators, \ie skill accuracy.
    This kind of fine-grained annotation is inspired by the recent study on the educational question genration~\cite{ghanem-etal-2022-question} and is used to evaluate the controllability of generation.
    Before the annotation, we show the annotators template samples for each comprehension skill as summarized in Table~\ref{tab:skill-schema}.
    During annotation, they are informed of the annotation instruction in the three steps. (1) Make a statement using the reference question and gold standard answer. (2) Extract sentences from the context required to support the statement. (3) Re-read our defined skill-based schema in Table~\ref{tab:skill-schema} and choose only one required skill to understand an entailment from extracted context to the statement.
    
    \item \textbf{Knowledge quality.} Since evaluating the overall quality of knowledge is challenging~\cite{heinzerling-inui-2021-language,west-etal-2022-symbolic}, this aspect checks the groundedness and relevance of our generated knowledge text to the given context. Specifically, we first show the annotators the input context, candidate question, answer text, and corresponding generated knowledge text. After that, we ask the annotator to answer two questions (``\textit{does the generated knowledge make sense}'' and ``\textit{is the generated knowledge relevant to the input context}''). Only our \texttt{SkillQG} and knowledge-augmented baselines are involved with this aspect of evaluation, and the annotation option is either \textit{yes} or \textit{no}.
\end{itemize}
As shown in Figure~\ref{fig:software}, we develop a web application to present and collect the human evaluation results automatically. This software can send the candidate samples to the annotators, guide them to evaluate samples from the aforementioned three dimensions and finally post the annotation results to our server. These results are based on the original collection of the dataset and will not violate the rights of individuals and groups. Based on the results, we report the human evaluation results in Section~\ref{subsec:main-results} and Section~\ref{sec:more-experiments}.

\begin{table}[!t]
    \centering
     \resizebox{0.7\columnwidth}{!}{
        \begin{tabular}{c|c|c}
            \toprule
            \textbf{Method}& \textbf{Makes Sense}& \textbf{Relevant} \\
            \midrule
            \midrule
            CsQG& 85.60\%&    85.00\% \\
            CQG& 77.70\%& 78.90\% \\
            SkillQG& 85.30\%& 90.40\% \\
            \bottomrule
        \end{tabular}
     }
    \caption{Human evaluation results on knowledge quality.
    We report the percentage of \emph{yes} answers for the two involved questions described in Section~\ref{subsec:evalution-prococol}
    .}
    \label{tab:knowledge-quality}
    \vspace{-5mm}
\end{table}

\section{More Experimental Results}
\label{sec:more-experiments}
We also analyze the quality of generated knowledge and better understand its contribution to the final performance.
The human evaluation results on the knowledge quality are summarized in Table~\ref{tab:knowledge-quality} and the inter-annotator Krippendorff's $\alpha$ is 88.42, indicating an acceptable level of consistency ($>$ 80\%) between annotators~\cite{krippendorff2004reliability}.
The few annotation conflicts are addressed after a discussion among the annotators.
The table shows that \texttt{SkillQG} can generate implicit knowledge that makes sense and is pertinent to the context for around 85\% of the time as evaluated by human annotators.
Compared with other knowledge-augmented baselines that retrieve knowledge from ConceptNet, \texttt{SkillQG} generates knowledge that is similar in terms of common sense and has better relevance to the input context.
The possible reason behind it is that \texttt{SkillQG} generates knowledge by asking and answering information-seeking questions based on the given context, benefiting the specialization of general knowledge of language models to each sample.


\end{document}